\RequirePackage{fix-cm}
\documentclass[smallextended]{svjour3}       

\usepackage{graphicx}
\usepackage{subfigure} 
\usepackage{booktabs}
\usepackage{multirow}
\usepackage{bm}
\usepackage{amsmath,amssymb,amsfonts}
\usepackage{algorithm}  
\usepackage{algpseudocode}  
\usepackage{amsmath}  
\usepackage{bbding}
\usepackage{pifont}
\usepackage{wasysym}
\usepackage{amssymb}
\usepackage{url}
\usepackage{hyperref}

\smartqed  
\usepackage{graphicx}
\begin{document}

\title{A Multi-task Joint Framework for Real-time Person Search
}


\author{Ye Li \and Kangning Yin \and Jie Liang \and Chunyu Wang \and Guangqiang Yin 
}


\institute{Ye Li \at
              Qingshuihe Campus of UESTC, No.2006, Xiyuan Avenue, West Hi-tech Zone,   Chengdu, Sichuan, P.R.China \\
              \email{liye@std.uestc.edu.cn}           
           \and
           S. Author \at
              second address
}

\date{Received: date / Accepted: date}

\maketitle

\begin{abstract}
Person search  generally involves three important parts: person detection, feature extraction and identity comparison. However, person search integrating detection, extraction and comparison has the following drawbacks. First, the accuracy of detection will affect the accuracy of comparison. Second, it is difficult to achieve real-time in real-world applications. To solve these problems, we propose a Multi-task Joint Framework for real-time person search (MJF), which optimizes the person detection, feature extraction and identity comparison respectively. For the person detection module, we proposed the YOLOv5-GS model, which is trained with person dataset. It combines the advantages of the Ghostnet and the Squeeze-and-Excitation (SE) block, and improves the speed and accuracy. For the feature extraction module,  we design the Model Adaptation Architecture (MAA), which could select different network according to the number of people. It could balance the relationship between accuracy and speed. For identity comparison, we propose a Three Dimension (3D) Pooled Table and a matching strategy to improve identification accuracy. On the condition of 1920*1080 resolution video and 500 IDs table, the identification rate (IR) and frames per second (FPS) achieved by our method could reach 93.6\% and 25.7, respectively. Therefore, the MJF could achieve the real-time person search.
\end{abstract}

\section{Introduction}
\label{intro}
Person search \cite{101} is mainly used for determining whether there is a specific person in an image or video sequence. Recently, it has attracted a lot of attention in the field of computer vision, especially in intelligent video surveillance and intelligent security.
In real-world application, person search can be divided into two tasks: person detection and person re-identification. The process is that detection is the premise of re-identification, and that neither can exist effectively without the other. However, most scholars only study person re-identification \cite{102,103,104}. Their achievements are difficult to be implemented in real-world applications. In addition, although some scholars propose an end-to-end network \cite{105,106,107} combined with two tasks, the accuracy is relatively low since the detection accuracy will affect the identification accuracy. Furthermore, person detection aims to distinguish the person from the background, while person identification aims to distinguish between different IDs. Therefore, a single network could not meet real-time requirement of person search. 

\begin{figure*}
	\includegraphics[width=0.99\textwidth]{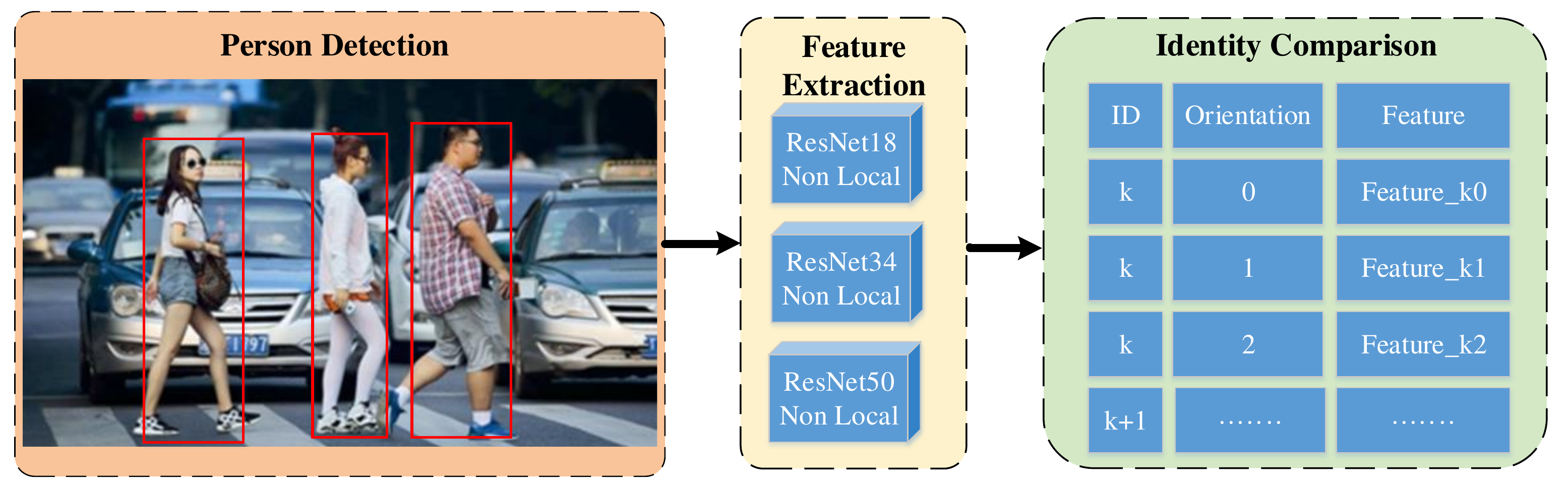}
	\caption{The Multi-task Joint Framework for real-time person search (MJF)}
	\label{framework}       
\end{figure*}

To address the problems above, we propose a Multi-task Joint Framework for real-time person search (MJF), which consists of the person detection, feature extraction and identity comparison, as shown in Fig.~\ref{framework}

Regarding the task of person detection, we integrate the Ghost \cite{108} and SE block \cite{109}, and propose the YOLOv5-GS network to improve the speed and accuracy. In addition, considering that the original YOLOv5 model is mainly used to detect multiple types of targets rather than single person, we train it with the CrowdHuman person dataset, which makes the network more suitable for person detection. As a result, our model could achieve 60.4 FPS on a 1920x1080 revolution video.

Regarding the task of feature extraction, we design a Model Adaptation Architecture (MAA), which could select different feature extraction models to balance the relationship between accuracy and speed. The MAA chooses simple model to extract feature when there are lots of people in a frame, while it chooses complex model to extract feature when there are a few people in a frame. Therefore, the module could improve accuracy as well as the speed. We choose ResNet-18, ResNet-34 and ResNet-50 networks \cite{110} as the basic models, and embed Non-local block \cite{111} in them. Considering that the features of a person are different in different orientations, we insert the person orientation classification module into the network, and extract orientation information in front, back and side respectively. The experiments show that the MAA could extract the features of 605 people in one second even if we use the ResNet-50 as the feature extraction model, and the mAP and Rank-1 can reach 93.2\% and 94.2\%, respectively.

Regarding the task of identity comparison, we design the 3D Pooled Table, which is used to store the features of each ID under three orientations. In addition, we propose the matching strategy that confirms 4 frames in 5 consecutive frames. After the confirmation, the feature is matched with the features in the 3D Pooled Table. If the match is successful, the feature under the corresponding ID and orientation will be updated; Otherwise, the new ID and orientation will be initialized. We use cosine distance to evaluate the similarity between different IDs. On the condition of 1920*1080 resolution video and 500 IDs table, the IR and FPS achieved by our method could reach 93.6\% and 25.7.

The main contributions of this paper are summarized as follows:
\begin{itemize}
	\item We optimize the person detection and propose the YOLOv5-GS model. It integrates Ghostnet and SE block, which reduce parameters and improve the detection accuracy;
	\item We propose a Model Adaptation Architecture (MAA). The architecture could flexibly select feature extraction networks according to the number of people and greatly improve the accuracy and speed. We insert the person orientation classification module into the network, and extract orientation information in front, back and side respectively;
	\item We design a 3D Pooled Table and propose the identity matching strategy. The stability and reliability of person identification are improved by identity confirming in 4 frames within 5 consecutive frames and feature matching in the same direction.
\end{itemize}

\section{Related Work}
\label{related_work}
\subsection{Person detection}
The object detection models can be divided into two categories. One is two-stage, which has two steps: object positioning and object recognition, such as R-CNN \cite{112}, Fast R-CNN \cite{113}, Faster R-CNN \cite{114}, etc. The candidate boxes are generated in the first step and the candidate boxes are judged in the second step. The generation and judgment of the boxes are two processes. This type of algorithms have high accuracy but slow speed. The other one is one-stage, such as SSD \cite{115}, YOLO \cite{116,117,118,119}, etc. The detection speed of this method is fast, which can meet the real-time requirements.

YOLO is the first end-to-end network, which could predict the locations and categories simultaneously. The YOLO series algorithms are divided into three modules: Backbone, Neck and Head. YOLO redefines object detection as a regression problem, which leads to high speed \cite{116,117}. Specially, the detection speed of YOLO is faster than other object detection algorithms for the simplification of network \cite{118} and uses of tricks \cite{119}, which can achieve real-time detection in the application of object detection \cite{120}.

\subsection{Feature extraction}
Feature extraction can be divided into global feature extraction and local feature extraction in the field of person re-identification. Generally, the input of global feature extraction network should be the whole images. Many scholars used ResNet, including ResNet-18, ResNet-34, ResNet-50, etc., which can solve the degradation problems. In addition, a series of networks have been developed to achieve batter performance, such as ResNeXt \cite{122} that combines ResNet and Inception \cite{121}, ResNest \cite{123} that fuses split-Attention module. However, the method of using global features has poor performance in real scenes for the problem of occlusion. Thus, the local feature extraction become extremely important. The global feature consists of multiple local features, which is extracted from a certain region. These methods include image segmentation \cite{124,125} and attitude extraction \cite{126,127}. Zhao et al. \cite{127} proposed a multi-stage feature decomposition and tree-like structure competitive feature fusion network, which is based on human body region guidance. It extracted people features with the help of 14 key position information of human body. To solve the problem of person dislocation, Zhao et al.\cite{128} proposed an effective person alignment network, which included convolutional network and region extraction network. It extracted the most discriminative human body areas and spliced them into the final person features.

\subsection{Identity comparison}
There are two key issues that need to be addressed in identity comparison. One is to confirm that the Tracklets in the scene belong to the same ID. The other one is to confirm that the ID is the comparison target. Xiao et al. \cite{129} proposed to use the LUT to store the features of each ID between the detection network and the person re-identification network. They trained an end-to-end person re-identification network using OIM loss function. Li et al. \cite{130} modified the LUT and proposed the Pooled Table (PT). It stored the features of each ID in different cameras. Considering the change of color, lighting and angle of person in different cameras, they proposed a new loss function Triplet Online Instance Matching Loss.

\section{Method}
\label{method}

\subsection{YOLOv5-GS}
It is difficult to achieve real-time person search by using the current methods since the detection is time-consuming. In this paper, we design the YOLOv5-GS model, which uses the idea of YOLOv4 [17] and the architecture of YOLOv5. YOLOv5 adopts the CSP[31] structure that can effectively reduce network parameters and the Focus \cite{131} structure that can decrease information loss, and the SPP pyramid structure that is suitable for multi-size input [31]. Inspired by these methods, we insert the Ghostnet and SE block into the YOLOv5 to improve the accuracy and speed. Considering that the reduction of parameters may affect the accuracy of detection, we add the SE layer to lower this effect. Although the insert of SE layer may increase the calculation, the model could strengthen valuable feature and suppress invaluable feature by adjust channel attention, which leads to the enhancement of learning ability of network. Specially, the person dataset CrowdHuman is used to train the YOLOv5-GS.
\begin{figure}
	\includegraphics[width=0.99\textwidth]{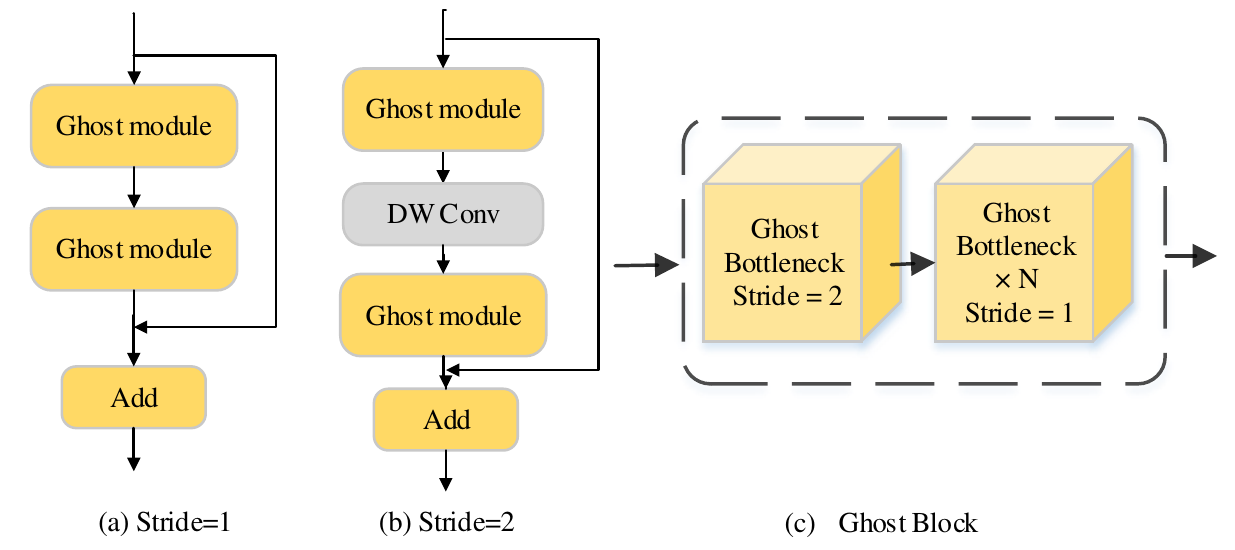}
	\caption{(a) The structure of GhostBottleneck \cite{108} when stride = 1. (b) The structure of GhostBottleneck when stride = 2. (c) The structure of Ghost Block.}
	\label{GhostBottleneck}       
\end{figure}

\begin{figure*}
	\includegraphics[width=0.99\textwidth]{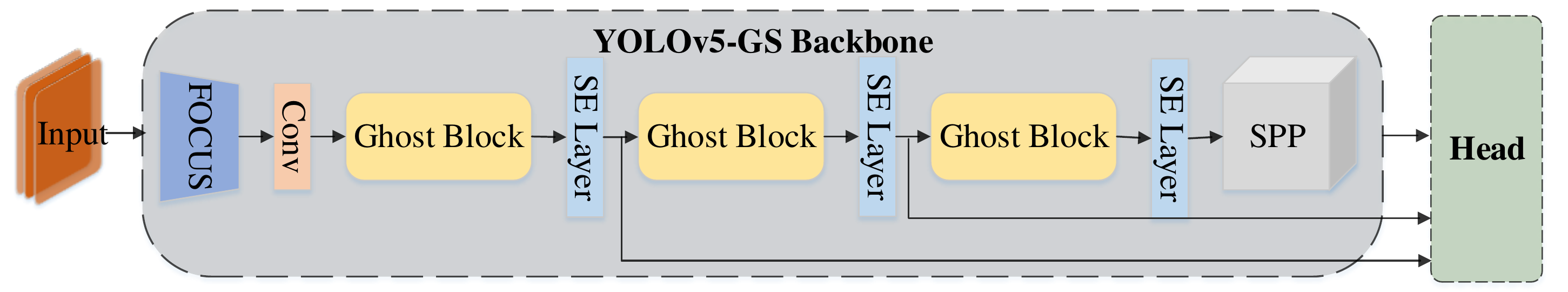}
	\caption{The architecture of Yolov5-GS}
	\label{fig3}       
\end{figure*}

\begin{figure*}
	\includegraphics[width=0.99\textwidth]{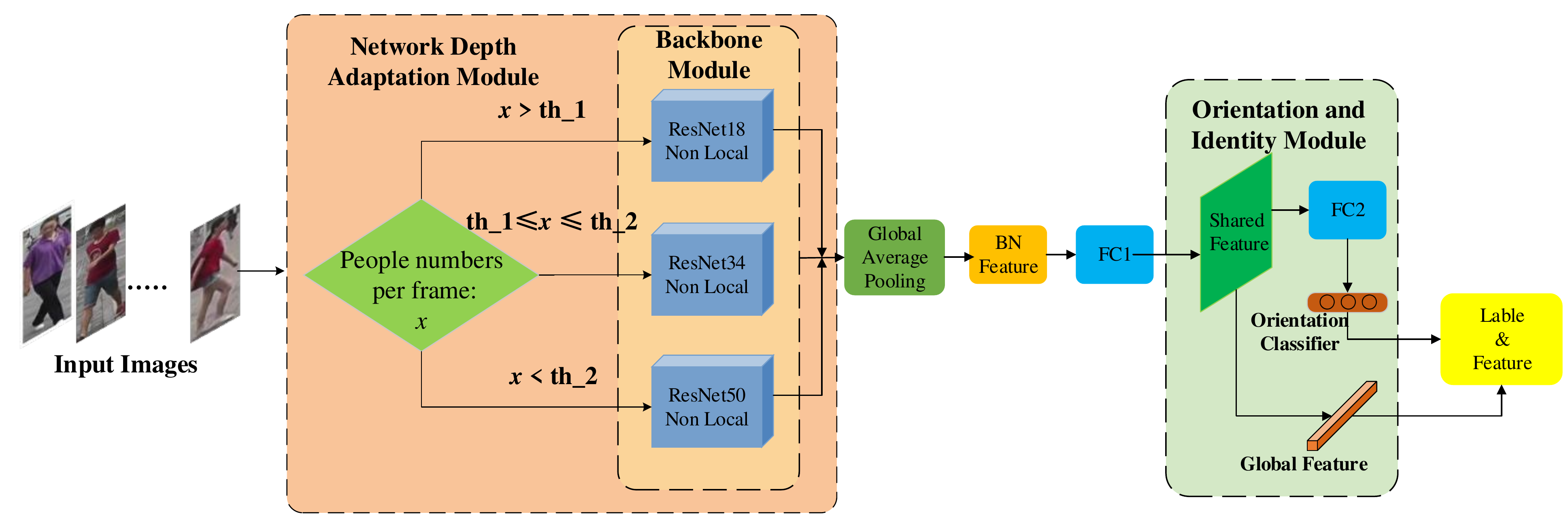}
	\caption{The architecture of MAA}
	\label{extraction_architgecture}       
\end{figure*}

The architecture of YOLOv5-GS is shown in Fig.~\ref{fig3}. We insert the Ghost Block into the backbone, N =1 in the first Ghost Block and the last two N=9. The detailed method is as follows: Fig.~\ref{GhostBottleneck} (a) as a residual Block replaces the CSP in the original structure and Fig.~\ref{GhostBottleneck} (b) replaces the convolutional layer, the two block make up the Ghost Block (as shown in Fig.~\ref{GhostBottleneck} (c)). We added a SE Layer after each sampling to enhance the feature learning ability.

\subsection{Model adaptation architecture (MAA)}

There are two problems needed to be considered in the feature extraction part. First, the features of the same ID are different under different orientations. If the orientation is ignored, the accuracy of person identification will decrease after changing orientation. Second, the speed of feature extraction decrease with the increase of layers. With the deepening of the network depth and the increase of structure complexity, the computing time will be extended, which affects the real-time performance. To solve the above problems, we design the MAA (As shown in Fig.\ref{extraction_architgecture}) which mainly consists of three parts: Backbone Module (BM); Network Depth Adaptation Module (NDAM); Orientation and Identity Module (OIM);

\paragraph{Backbone module (BM).} In this section, we mainly study the three backbone networks: ResNet-18, ResNet-34 and ResNet-50. The structure of the BM is shown in Fig.\ref{extraction_architgecture}. We use Non-local block to establish the connection between the pixels that have a certain distance and Bn Feature [32] structure to classify orientation and feature. The detailed network structure is shown in Table \ref{network}.

\begin{table}[!htbp]
	\centering
	\caption{The structure of Backbone Module}
	\setlength{\tabcolsep}{3.3mm}{
		\begin{tabular}{ccccccc}
			&\multicolumn{2}{c}{ResNet18}&\multicolumn{2}{c}{ResNet-34}&\multicolumn{2}{c}{ResNet-50} \\
			\toprule
			\textit{conv}$_1$&  \multicolumn{3}{c}{$7\times7, 64$, stride $2$} & $\times1$&   \\
			\midrule
			\textit{pool}$_1$&	\multicolumn{3}{c}{$3\times3$, avg, stride $2$}&$\times1$& \\
			\midrule
			&$3\times3$, 64& & $3\times3$, 64& & $1\times1$, 64& \\
			\textit{layer}$_1$     & & $\times2$& &$\times3$ & $3\times3$, 64&$\times3$\\
			&$3\times3$, 64& & $3\times3$, 64& & $1\times1$, 64&\\
			\midrule
			\textit{Non-local}$_1$     &$1\times1$, 64 & $\times1$& $1\times0$, 64&$\times0$ & $1\times1$, 256&$\times0$\\
			\midrule
			&$3\times3$, 128& & $3\times3$, 128& & $1\times1$, 128& \\
			\textit{layer}$_2$    & & $\times2$& &$\times4$ & $3\times3$, 128&$\times4$\\
			&$3\times3$, 128& & $3\times3$, 128& & $1\times1$, 512&\\
			\midrule
			\textit{Non-local}$_2$     &$1\times1$, 128 & $\times1$& $1\times1$, 256&$\times1$ & $1\times1$, 1024&$\times1$\\		
			\midrule	
			&$3\times3$, 256& & $3\times3$, 256& & $1\times1$, 256& \\
			\textit{layer}$_3$    & & $\times2$& &$\times6$ & $3\times3$, 256&$\times6$\\
			&$3\times3$, 256& & $3\times3$, 256& & $1\times1$, 1024&\\
			\midrule
			\textit{Non-local}$_3$     &$1\times1$, 256 & $\times1$& $1\times1$, 256&$\times1$ & $1\times1$, 1024&$\times0$\\	
			\midrule
			&$3\times3$, 512& & $3\times3$, 512& & $1\times1$, 512& \\
			\textit{layer}$_4$    & & $\times2$& &$\times3$ & $3\times3$, 512&$\times3$\\
			&$3\times3$, 512& & $3\times3$, 512& & $1\times1$, 2048&\\
			\midrule
			\textit{pool}$_2$ & \multicolumn{6}{c}{global average pooling}\\
			\midrule
			\textit{fc}$_1$ & -& & -& & ($2048,512$) & $\times1$\\
			\midrule
			\textit{BN} &  \multicolumn{6}{c}{Bn Feature}\\
			\midrule
			\textit{fc}$_2$ & \multicolumn{6}{c}{$(512,3)$}\\
			\bottomrule
	\end{tabular}}
	\label{network}
\end{table}

\paragraph{Network Depth Adaptation Module (NDAM).} There will be a game process between accuracy and speed. The deeper the network is, the higher the recognition accuracy and the lower PPS will be. On the contrary, the shallower the network depth, the lower the recognition accuracy and the higher the PPS will be. Therefore, the feature extraction module should achieve balance to ensure high accuracy and high speed. We design the NDAM, as shown in Fig.~\ref{extraction_architgecture}. It can select the corresponding network according to the number of targets in each frame, and improve the efficiency of feature extraction without decreasing the accuracy. ResNet18 + Non-local + BnFeature (RN18), ResNet34 + Non-local + BnFeature (RN34) and ResNet50 + Non-local + BnFeature (RN50) are selected as the three adaptive branch networks. The formulate is expressed as follows. 

\begin{equation}
backbone =\left\{\begin{matrix}
	RN50 & n\leqslant th_{1}\\ 
	RN34 & th_{1}< n< th_{2} \\ 
	RN18 & n\geqslant th_{2}
\end{matrix}\right.
\end{equation}

where $n$ is the number of people in one frame, $th_{1}$ and $th_{2}$ represent the thresholds of people number. 

\paragraph{Orientation and identity module (OIM).} To reduce the influence of orientations, we proposed the OIM, which contains orientation feature and ID feature, as shown in Fig.\ref{extraction_architgecture}. We insert FC2 layer into the OIM to extract orientation features. We classify the orientations as front, back, and side. The person ID feature is represented by a vector of 512 dimensions, and the person orientation feature is represented by a vector of 3 dimensions, with a total of 515 dimensions.

\subsection{3D Pooled Table and matching strategy}
\begin{figure*}
	\includegraphics[width=0.99\textwidth]{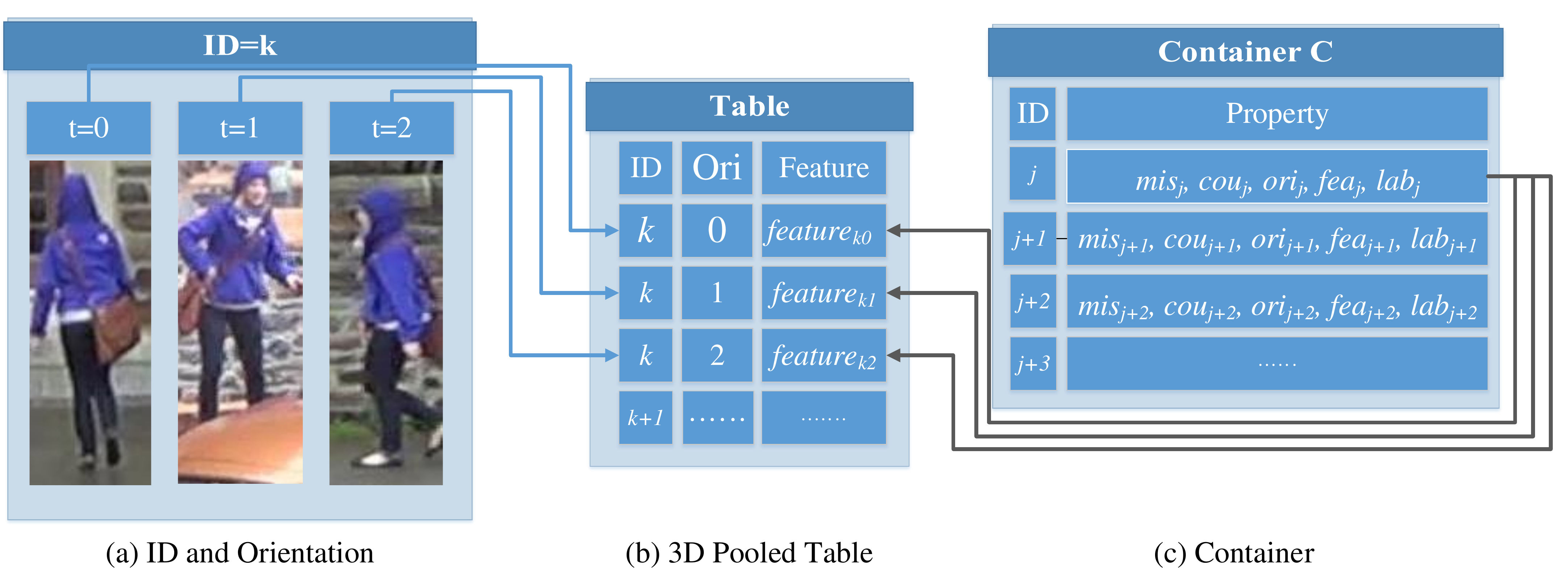}
	\caption{The process of feature storage and comparison. (a) ID and Orientation. (b) 3D Pooled Table. (c) Container. Person features of the same ID with different orientations will be stored in the 3D Pooled Table. The Container updates the state after processing each frame. After the match is confirmed, the feature in the container will replace the feature of corresponding ID and orientation in 3D Pooled Table.}
	\label{3D-pooled-table}       
\end{figure*}

In this section, a 3D Pooled Table is designed to store features of each ID in different orientations, as shown in Fig.\ref{3D-pooled-table}. Each feature is represented by $f\in R^{D}$, where $D$ represents the dimension of the feature, and the orientation is represented by $t\in R^{1}$, $t=0$ represents the front, $t=1$ represents the back, and $t=2$ represents the side. Each ID in the Table is represented by $V\in R^{D\times T}$, where $T=3$ represents the number of types of orientation; In addition, we design a matching strategy to confirm the same ID through container storage. The features and orientations of different ids will be stored in different containers, and the containers also contain the state of feature matching. If a container matches 4 frames in 5 consecutive frames, the state of the container will be changed to confirm, and the features in the container will be compared with the person features in the 3D Pooled Table with the same orientation. The detailed processes are as follows.

%
%
%

First, the 1st frame input the model and the fragments $p_{1}^{i}$ of all person in this frame is detected through YOLOv5-GS, where $i$ represents the $i$-$th$ person in the 1st frame. Each detected person will create a container $C_{x},x=1,2,3...$, where x represents the $x$-$th$ container. In each container, the corresponding person features will be stored, including the number of matches ($cou$) , the number of mismatches ($mis$), the features ($fea$), the orientation of pedestrians ($ori$), and the person label ($lab$). Then, the detected fragments of person input MAA for feature and orientation extraction, and store the features and orientations to the corresponding container $C_{x}$. When the $a$-$th$ frame input the model, the fragments $p_{a}^{i}$ of all the pedestrians in this frame are first extracted through YOLOv5-GS. The detected fragments of person input to network MAA to extract the feature $p_{a}^{i}\in R^{D}$ and the orientation $t_{a}^{i}\in R^{1}$. The extracted feature $f_{a}^{i}\in R^{D}$ is matched with the existing feature in the container, and if matched, the corresponding $fea$ and $cou$ in the container are updated. For containers that do not match, only the $mis$ is updated, and if the $mis$ is updated twice in 5 frames, the container is deleted. If a container matches 4 frames in 5 consecutive frames, then remove the container and match the features of the container with the corresponding orientation in the table $f_{n}^{t}\in R^{D}$, where $n$ represents the $n$-$th$ ID and $t$ represents the orientation. If the match is successful, it updates the features of the corresponding ID orientation in the table. Otherwise, the new ID will be initialized.

\section{Dataset and evaluation index}
\label{dataset}

\subsection{Dataset of person detection}
\paragraph{COCO} \cite{132}. It is a large image dataset designed for object detection, segmentation, person keypoints detection, stuff segmentation, and caption generation.
\paragraph{CrowdHuman}  \cite{133}. It is a benchmark dataset to better evaluate detectors in crowd scenarios. The CrowdHuman dataset is large, rich-annotated and contains high diversity. CrowdHuman contains 15000, 4370 and 5000 images for training, validation, and testing, respectively. 
\paragraph{TownCentreXVID} \cite{134}. The resolution is 1920*1080 and the video frame rate is 25 fps. The peak number of person in the video frame is 30. The average number of people in each frame is 16, with a total of 7,500 frames. It is a relatively complex person video dataset at present.
\paragraph{Evaluation index.} MAP, FPS, Positive Detection Rate (PDR) and Miss Detection Rate (MDR) are used to evaluate the performance of detection model.

\subsection{Dataset of feature extraction}
\paragraph{Market-1501} \cite{135}. The dataset is collected by 6 different cameras. The whole dataset contains 32,668 images, including 12,936 for training and 19,732 for testing.
\paragraph{DukeOrientation.} Images with different orientations (front, back and side) were selected from the DukeMTMC dataset \cite{136}. The DukeOrientation dataset contains 3587 images, including front 1,140, back 873 and side 1,574.
\paragraph{Evaluation index.} We use Rank-1, mAP and PPS to evaluate the performance of feature extraction. Person Per Second (PPS) refers to the number of pedestrians of feature extraction within one second.

\subsection{Dataset of real-time comparison}
\paragraph{TownCentreXVID.} In this part, we need to evaluate the performance of the whole framework. We adopt TownCentreXVID video, which has been described in 4.1. This dataset has many person targets and the background is complex, thus it can greatly reflect the performance of our framework.
\paragraph{Evaluation index.} FPS is used to evaluate the detection speed of the whole process and Identification Rate (IR) is used to evaluate the accuracy. The number of correct identification represents the number of correct identification in the identity comparison process and the total number represents the total number of identity comparisons.
\begin{equation}
IR = \frac{n_{cor}}{n_{all}}
\end{equation}
Where $n_{cor}$ is number of correct identification, $n_{all}$ is the total number of identification.

\section{Experiments}
\label{experiment}
\subsection{The comparison of person detection}
In this section, we first compare the performance of YOLOv5, YOLOv5-G (YOLOv5 + Ghostnet) and YOLOv5-GS (YOLOv5 + Ghostnet + SE layer) in the field of object detection on the COCO dataset. Then we select YOLOv5-GS as the person detection model. In addition, we use the CrowdHuman dataset train and test the model.

\paragraph{The comparison of different model in object detection.} We test the performance of YOLOv5, YOLOv5-G and YOLOv5-GS on COCO dataset. We compare the parameters, mAP and FPS. The experimental results are shown in Table \ref{detection}.

Compared with YOLOv5 network, the parameters of YOLOv5-GS decrease from 7.25M to 4.47M. The detection speed increases by 51 frames per second. At the same time, the results of YOLOv5-G and YOLOv5-GS demonstrate that SE module could compensate for the decline of mAP caused by the Ghostnet. Based on the above experiments, we adopted YOLOv5-GS as person detection model.

\paragraph{The effectiveness of YOLOv5-GS in person detection.} The model, which is trained with COCO dataset, is not suitable for person detection. A large number of non-person targets consume the detection time. We re-train the Yolov5-GS model with the CrowdHuman dataset. The validation video we adopted is TownCentreXVID[1] and we test the relation between FPS and resolution. The experimental results are shown in Table \ref{yolov5-gs}.

\begin{table}
	\footnotesize
	\centering
	\caption{Comparisons between the YOLOv5-GS and others on COCO dataset}
	\setlength{\tabcolsep}{2.9mm}{
		\begin{tabular}{cccc}
			\toprule
			Network&Parameter&mAP&FPS \\
			\midrule
			YOLOv5  & 7.25 M & 56.2 & 333 \\
			YOLOv5-G & 3.97 M& 49.8 & 400\\
			\textbf{YOLOv5-GS}&\textbf{4.47 M}& \textbf{53.1} &\textbf{384}\\
			\bottomrule
	\end{tabular}}
	\label{detection}
\end{table}

\begin{table}
	\footnotesize
	\centering
	\caption{The performance of accuracy and speed on different resolution}
	\setlength{\tabcolsep}{2.9mm}{
		\begin{tabular}{cccccc}
			\toprule
			Network&Resolution (original)&Resolution (resize)&FPS& PDR& MDR  \\
			\midrule
			\multirow{3}*{YOLOv5}  & \multirow{3}*{1920*1080} & 640*640 & 63.5 &95.5&8.7\\
			&  & 1088*1088 & 48.4 &98.1&5.9\\
			&  & 1920*1920 & 48.2 &99.3&4.5\\
			\midrule
			\multirow{3}*{YOLOv5-GS}  & \multirow{3}*{1920*1080} & 640*640 & 48.6 &96.6&8.4\\
			  &  & 1088*1088 & 48.4 &98.6&2.1\\
			  &  & 1920*1920 & 48.2 &99.5&2.5\\
			\bottomrule
	\end{tabular}}
	\label{yolov5-gs}
\end{table}

The results show that the detection speed of YOLOv5-GS could reach 60.4 FPS in 1920*1080 resolution video, which could meet the real-time application requirements. In addition, with the increase of the resolution of the input image, the speed of the network decreases and the detection accuracy increases. The experiment also proves that YOLOv5-GS can improve the detection speed while maintaining the detection accuracy.

\subsection{The comparison of feature extraction} 
The size of the input images is set to 128*64. We train the model on Market1501 dataset and use some tricks. The output dimensions is 512, and we test the mAP and Rank-1 to evaluate network accuracy. The PPS is used to evaluate extraction speed. The results are shown in Table \ref{different model}.

It can be seen from Table \ref{different model} that with the addition of Non-local and BnFeature, mAP will increase. Compared with Non-local, BnFeature has greater influence on mAP. Non-local increases the depth of the network and reduces the PPS, while the BnFeature has little influence on PPS. Based on the above results, ResNet18 + Non-local + BnFeature (RN18), ResNet34 + Non-local + BnFeature (RN34) and ResNet50 + Non-local + BnFeature (RN50) are selected as the three adaptive branch networks. According to the results in Table 4, we calculate the corresponding relationship between the number of people and the selection of feature extraction network. The selection strategy is shown in Table 5.

\begin{table}
	\footnotesize
	\centering
	\caption{The performance of different models}
	\setlength{\tabcolsep}{2.9mm}{
		\begin{tabular}{cccccc}
			\toprule
			\multirow{2}*{Network}&\multicolumn{2}{c}{Tricks}&	\multirow{2}*{mAP}&	\multirow{2}*{Rank-1}& 	\multirow{2}*{PPS}  \\
			\cmidrule{2-3}
			& Non-local & Bn Feature & & & \\
			\midrule
			\multirow{4}*{ResNet-18}  & \ding{53} & \ding{53} & 89.2 & 90.4 & 825.979\\ 
			 & \ding{51} &  \ding{53} & 89.7 & 90.8 & 723.284\\
			 & \ding{53} & \ding{51} & 91.1 & 92.5 & 795.735\\
			  & \ding{51} & \ding{51} & 91.8 & 93.1 & 709.321\\
			\midrule
			\multirow{4}*{ResNet-34}  & \ding{53} & \ding{53} & 89.4 & 90.7 & 717.265\\ 
			  & \ding{51} & \ding{53} & 90.7 & 92.2 & 664.488\\
              &\ding{53} & \ding{51} & 90.5 & 91.6 & 726.795\\
              & \ding{51} & \ding{51} & 93.2 & 94.2 & 637.340\\
             \midrule
             \multirow{4}*{ResNet-50}  & \ding{53} & \ding{53} & 89.3 & 91.2 & 625.607\\ 
              & \ding{51} & \ding{53} & 90.1 & 91.4 & 616.927\\
              & \ding{53} & \ding{51} & 93.2 & 94.2 & 624.981\\
              & \ding{51} & \ding{51} & 93.2 & 94.9 & 605.556\\
			\bottomrule
	\end{tabular}}
	\label{different model}
\end{table}

According to Table \ref{threshold}, RN50 is selected when the number of pedestrians in the frame is less than 24; RN34 is selected when the number of pedestrians in the frame is less than 25; and RN18 is selected when the number of pedestrians in the frame is less than 28.

\begin{table}
	\footnotesize
	\centering
	\caption{Threshold table}
	\setlength{\tabcolsep}{3.9mm}{
		\begin{tabular}{ccccc}
			\toprule
			Network&PPS&FPS&Max number\\
			\midrule
			RN18  & 709.321 & 25 & 28.37 \\
			RN34 & 637.340 & 25 & 25.49\\
			RN50 & 605.556 & 25 & 24.22\\
			\bottomrule
	\end{tabular}}
	\label{threshold}
\end{table}

\subsection{The comparison of identity}
In this section, we study the impact of resolution and the number of IDs in the 3D Pooled Table on identification rate and FPS. The resolution will affect the accuracy and speed of detection, and the number of IDs in the 3D Pooled Table will affect the accuracy and speed of comparison. The experimental results are shown in Table \ref{comparison}.

\begin{table}
	\footnotesize
	\centering
	\caption{The influence of video resolution and ID number on the accuracy and speed}
	\setlength{\tabcolsep}{3.9mm}{
		\begin{tabular}{ccccccccc}
			\toprule
			\multirow{2}*{IDs}&\multicolumn{2}{c}{1920*1080}& &\multicolumn{2}{c}{1280*720}& &\multicolumn{2}{c}{720*480} \\
			\cmidrule{2-3}
			\cmidrule{5-6}
			\cmidrule{8-9}
			&IR&FPS& & IR&FPS & &IR&FPS \\
			\midrule
			100 & 97.6&27.9& &96.2&28.6& &95.4&33.5\\
			200& 96.9&26.8& &96.1&28.6& &95.6&32.9\\
			300& 95.0&26.8& &94.6&28.1& &94.1&31.5\\
			400& 94.8&25.5& &94.8&27.9& &94.1&31.8\\
			500& 93.6&25.7& &92.9&27.5& &92.5&29.9\\
			\bottomrule
	\end{tabular}}
	\label{comparison}
\end{table}

It can be seen from Table 6 that the increase of the number of IDs results in the decrease of IR and FPS. This is because the more IDs, the more person will need to be compared. In addition, the IR decreases with the decrease of resolution, but the comparison speed is improved to some extent. The main reason is that the poor feature extracted from low resolution pedestrian images leads to the decrease of IR. However, this kind of images improve the speed of detection. In conclusion, On the condition of 1920*1080 resolution video and 500 ID table, the identification rate (IR) and frames per second (FPS) achieved by our method could reach 93.6\% and 25.7, respectively.

\section{Conclusion}
In this paper, we propose a multi-task joint framework for real-time person search, which integrates person detection, feature extraction and identity comparison. The accuracy and speed of each part are optimized respectively to achieve real-time. On the condition of 1920*1080 resolution video and 500 IDs table, the identification rate (IR) and frames per second (FPS) achieved by our method could reach 93.6\% and 25.7, respectively. It could meet the real-time application requirements. We also provide a reference solution for real-time person search.

\paragraph{Declarations}

The data that support the findings of this study are available online. These datasets were derived from the following public domain resources:
[\href{http://cocodataset.org}{COCO}, \href{http://www.crowdhuman.org/}{CrowdHuman}, \href{https://drive.google.com/file/d/0B8-rUzbwVRk0c054eEozWG9COHM/view}{Market-1501},  \href{https://drive.google.com/file/d/1jjE85dRCMOgRtvJ5RQV9-Afs-2_5dY3O/view}{ DukeMTMC}, \href{https://www.youtube.com/watch?v=rfkGy6dwWJs}{TownCentreXVID}]




%


%
%

\bibliographystyle{spmpsci}      

\bibliography{cvprbib}

\begin{thebibliography}{10}
\providecommand{\url}[1]{{#1}}
\providecommand{\urlprefix}{URL }
\expandafter\ifx\csname urlstyle\endcsname\relax
  \providecommand{\doi}[1]{DOI~\discretionary{}{}{}#1}\else
  \providecommand{\doi}{DOI~\discretionary{}{}{}\begingroup
  \urlstyle{rm}\Url}\fi

\bibitem{134}
Benfold, B., Reid, I.: Stable multi-target tracking in real-time surveillance
  video.
\newblock pp. 3457 -- 3464 (2011).
\newblock \doi{10.1109/CVPR.2011.5995667}

\bibitem{119}
Bochkovskiy, A., Wang, C.Y., Liao, H.Y.M.: Yolov4: Optimal speed and accuracy
  of object detection.
\newblock arXiv preprint arXiv:2004.10934  (2020)

\bibitem{118}
Farhadi, A., Redmon, J.: Yolov3: An incremental improvement.
\newblock Computer Vision and Pattern Recognition, cite as  (2018)

\bibitem{113}
Girshick, R.: Fast r-cnn.
\newblock In: Proceedings of the IEEE international conference on computer
  vision, pp. 1440--1448 (2015)

\bibitem{112}
Girshick, R., Donahue, J., Darrell, T., Malik, J.: Rich feature hierarchies for
  accurate object detection and semantic segmentation.
\newblock In: Proceedings of the IEEE conference on computer vision and pattern
  recognition, pp. 580--587 (2014)

\bibitem{136}
{Gou}, M., {Karanam}, S., {Liu}, W., {Camps}, O., {Radke}, R.J.: Dukemtmc4reid:
  A large-scale multi-camera person re-identification dataset.
\newblock In: 2017 IEEE Conference on Computer Vision and Pattern Recognition
  Workshops (CVPRW), pp. 1425--1434 (2017).
\newblock \doi{10.1109/CVPRW.2017.185}

\bibitem{108}
Han, K., Wang, Y., Tian, Q., Guo, J., Xu, C., Xu, C.: Ghostnet: More features
  from cheap operations.
\newblock In: Proceedings of the IEEE/CVF Conference on Computer Vision and
  Pattern Recognition, pp. 1580--1589 (2020)

\bibitem{110}
He, K., Zhang, X., Ren, S., Sun, J.: Deep residual learning for image
  recognition.
\newblock In: Proceedings of the IEEE conference on computer vision and pattern
  recognition, pp. 770--778 (2016)

\bibitem{104}
He, T., Zhang, Z., Zhang, H., Zhang, Z., Xie, J., Li, M.: Bag of tricks for
  image classification with convolutional neural networks.
\newblock In: Proceedings of the IEEE Conference on Computer Vision and Pattern
  Recognition, pp. 558--567 (2019)

\bibitem{107}
He, Z., Zhang, L.: End-to-end detection and re-identification integrated net
  for person search.
\newblock In: Asian Conference on Computer Vision, pp. 349--364. Springer
  (2018)

\bibitem{120}
Hu, J., Gao, X., Wu, H., Gao, S.: Detection of workers without the helments in
  videos based on yolo v3.
\newblock In: 2019 12th International Congress on Image and Signal Processing,
  BioMedical Engineering and Informatics (CISP-BMEI), pp. 1--4. IEEE (2019)

\bibitem{109}
Hu, J., Shen, L., Sun, G.: Squeeze-and-excitation networks.
\newblock In: Proceedings of the IEEE conference on computer vision and pattern
  recognition, pp. 7132--7141 (2018)

\bibitem{124}
Li, W., Zhao, R., Xiao, T., Wang, X.: Deepreid: Deep filter pairing neural
  network for person re-identification.
\newblock In: Proceedings of the IEEE conference on computer vision and pattern
  recognition, pp. 152--159 (2014)

\bibitem{130}
Li, Y., Yin, G., Liu, C., Yang, X., Wang, Z.: Triplet online instance matching
  loss for person re-identification.
\newblock arXiv preprint arXiv:2002.10560  (2020)

\bibitem{132}
Lin, T.Y., Maire, M., Belongie, S., Hays, J., Perona, P., Ramanan, D.,
  Doll{\'a}r, P., Zitnick, C.L.: Microsoft coco: Common objects in context.
\newblock In: European conference on computer vision, pp. 740--755 (2014)

\bibitem{115}
Liu, W., Anguelov, D., Erhan, D., Szegedy, C., Reed, S., Fu, C.Y., Berg, A.C.:
  Ssd: Single shot multibox detector.
\newblock In: European conference on computer vision, pp. 21--37. Springer
  (2016)

\bibitem{106}
Munjal, B., Amin, S., Tombari, F., Galasso, F.: Query-guided end-to-end person
  search.
\newblock In: Proceedings of the IEEE Conference on Computer Vision and Pattern
  Recognition, pp. 811--820 (2019)

\bibitem{116}
Redmon, J., Divvala, S., Girshick, R., Farhadi, A.: You only look once:
  Unified, real-time object detection.
\newblock In: Proceedings of the IEEE conference on computer vision and pattern
  recognition, pp. 779--788 (2016)

\bibitem{117}
Redmon, J., Farhadi, A.: Yolo9000: better, faster, stronger.
\newblock In: Proceedings of the IEEE conference on computer vision and pattern
  recognition, pp. 7263--7271 (2017)

\bibitem{114}
Ren, S., He, K., Girshick, R., Sun, J.: Faster r-cnn: Towards real-time object
  detection with region proposal networks.
\newblock In: Advances in neural information processing systems, pp. 91--99
  (2015)

\bibitem{133}
Shao, S., Zhao, Z., Li, B., Xiao, T., Yu, G., Zhang, X., Sun, J.: Crowdhuman: A
  benchmark for detecting human in a crowd.
\newblock arXiv preprint arXiv:1805.00123  (2018)

\bibitem{126}
Su, C., Li, J., Zhang, S., Xing, J., Gao, W., Tian, Q.: Pose-driven deep
  convolutional model for person re-identification.
\newblock In: Proceedings of the IEEE international conference on computer
  vision, pp. 3960--3969 (2017)

\bibitem{103}
Sun, Y., Zheng, L., Deng, W., Wang, S.: Svdnet for pedestrian retrieval.
\newblock In: Proceedings of the IEEE International Conference on Computer
  Vision, pp. 3800--3808 (2017)

\bibitem{121}
Szegedy, C., Liu, W., Jia, Y., Sermanet, P., Reed, S., Anguelov, D., Erhan, D.,
  Vanhoucke, V., Rabinovich, A.: Going deeper with convolutions.
\newblock In: Proceedings of the IEEE conference on computer vision and pattern
  recognition, pp. 1--9 (2015)

\bibitem{125}
Varior, R.R., Shuai, B., Lu, J., Xu, D., Wang, G.: A siamese long short-term
  memory architecture for human re-identification.
\newblock In: European conference on computer vision, pp. 135--153. Springer
  (2016)

\bibitem{131}
Wang, C.Y., Mark~Liao, H.Y., Wu, Y.H., Chen, P.Y., Hsieh, J.W., Yeh, I.H.:
  Cspnet: A new backbone that can enhance learning capability of cnn.
\newblock In: Proceedings of the IEEE/CVF Conference on Computer Vision and
  Pattern Recognition Workshops, pp. 390--391 (2020)

\bibitem{111}
Wang, X., Girshick, R., Gupta, A., He, K.: Non-local neural networks.
\newblock In: Proceedings of the IEEE conference on computer vision and pattern
  recognition, pp. 7794--7803 (2018)

\bibitem{105}
Xiao, T., Li, S., Wang, B., Lin, L., Wang, X.: End-to-end deep learning for
  person search.
\newblock arXiv preprint arXiv:1604.01850 \textbf{2}(2) (2016)

\bibitem{129}
Xiao, T., Li, S., Wang, B., Lin, L., Wang, X.: Joint detection and
  identification feature learning for person search.
\newblock In: Proceedings of the IEEE Conference on Computer Vision and Pattern
  Recognition, pp. 3415--3424 (2017)

\bibitem{122}
Xie, S., Girshick, R., Doll{\'a}r, P., Tu, Z., He, K.: Aggregated residual
  transformations for deep neural networks.
\newblock In: Proceedings of the IEEE conference on computer vision and pattern
  recognition, pp. 1492--1500 (2017)

\bibitem{101}
Xu, Y., Ma, B., Huang, R., Lin, L.: Person search in a scene by jointly
  modeling people commonness and person uniqueness.
\newblock In: Proceedings of the 22nd ACM international conference on
  Multimedia, pp. 937--940 (2014)

\bibitem{123}
Zhang, H., Wu, C., Zhang, Z., Zhu, Y., Zhang, Z., Lin, H., Sun, Y., He, T.,
  Mueller, J., Manmatha, R., et~al.: Resnest: Split-attention networks.
\newblock arXiv preprint arXiv:2004.08955  (2020)

\bibitem{127}
Zhao, H., Tian, M., Sun, S., Shao, J., Yan, J., Yi, S., Wang, X., Tang, X.:
  Spindle net: Person re-identification with human body region guided feature
  decomposition and fusion.
\newblock In: Proceedings of the IEEE conference on computer vision and pattern
  recognition, pp. 1077--1085 (2017)

\bibitem{128}
Zhao, L., Li, X., Zhuang, Y., Wang, J.: Deeply-learned part-aligned
  representations for person re-identification.
\newblock In: Proceedings of the IEEE international conference on computer
  vision, pp. 3219--3228 (2017)

\bibitem{135}
Zheng, L., Shen, L., Tian, L., Wang, S., Wang, J., Tian, Q.: Scalable person
  re-identification: A benchmark.
\newblock In: Proceedings of the IEEE international conference on computer
  vision, pp. 1116--1124 (2015)

\bibitem{102}
Zhong, Z., Zheng, L., Cao, D., Li, S.: Re-ranking person re-identification with
  k-reciprocal encoding.
\newblock In: Proceedings of the IEEE Conference on Computer Vision and Pattern
  Recognition, pp. 1318--1327 (2017)

\end{thebibliography}


\end{document}